\documentclass{llncs}
\usepackage{url}

\title{Proofs, proofs, proofs, and proofs\thanks{The final publication of this paper is available at www.springerlink.com}}

\author{Manfred Kerber}
\institute{Computer Science,
  University of Birmingham \\
  Birmingham B15 2TT, England \\
  {\tt http://www.cs.bham.ac.uk/\~{}mmk}} 

\begin{document}
\maketitle

\begin{abstract}
  In logic there is a clear concept of what constitutes a proof and
  what not. A proof is essentially defined as a finite sequence of
  formulae which are either axioms or derived by proof rules from
  formulae earlier in the sequence. Sociologically, however, it is
  more difficult to say what should constitute a proof and what not.
  In this paper we will look at different forms of proofs and try to
  clarify the concept of proof in the wider meaning of the term. This has
  implications on how proofs should be represented formally.
\end{abstract}

\section{Introduction}\label{sec:intro}

There is a relatively clear definition of what a \textit{proof\/}
is. The concept has been clarified in the last 150 years with the
development of logic and in the last 50 to 60 years with the
development of systems which formalize the results of these
investigations in formal computer systems. Mathematicians, however,
seem not to have much changed their view of proofs.\footnote{Obviously
  for those working on the foundations of mathematics this is
  different. The generalization `mathematicians' does not mean all
  mathematicians, but most typical mathematicians.}  Sure, they have
some knowledge of the results in foundations but if they work in
fields such as statistics, group theory, or geometry then the
formalization of proof is only of marginal interest, although the
concept of proof itself is at the core of the whole of mathematics.

Is this just a matter of ignorance? Or rather of professionalism? And
what are the consequences for our field which tries to offer support
for mathematicians?

In order to approach these questions an account of the development of
the concept of proof in different fields is given. We first take a
look at the development in logic
(section~\ref{sec:logic})\footnote{The claim is not that there is a
  single view in the different fields. Even a single person may have
  different views at different times or in different contexts.}. Next
we see the consequences this had on working mathematicians and their
attitude towards formal proofs (section~\ref{sec:maths}). The
development in logic has strongly influenced the development of
deduction systems. In section~\ref{sec:sys} we take a brief look at
deduction systems. Then some consequences for the field of
mathematical knowledge management are discussed. Essentially we argue
that the representation of proofs should be flexible enough to serve
different purposes in order to be able to communicate proofs at
different levels: checkable proofs, abstract high-level proofs, proof
ideas/plans, and false proofs.

\section{The Logician's View}\label{sec:logic}

In the second half of the 19th century and in the early 20th century,
a rigorous definition of the concept of proof was given. Inspired by
the rigorous work of Euclid, Hilbert axiomatized geometry and
developed a programme to be carried through for all of
mathematics. Whitehead and Russell wrote the Principia
Mathematica~\cite{Whitehead10} which started to implement the grand
vision to reduce all of mathematics to logic. Logicians like Boole and
Frege developed propositional and predicate logic, and Gentzen the
calculus of Natural Deduction. 

As Frege put it, the vision was (\cite{Frege79} quoted
from~\cite[p.6f]{Heijenoort67}):
\begin{quote}\sl
In apprehending a scientific truth we pass, as a rule, through
various degrees of certitude. Perhaps first conjectured on the basis of
an insufficient number of particular cases, a general proposition
comes to be more and more securely established by being connected with
other truths through chains of inferences, whether consequences are
derived from it that are confirmed in some other way or whether,
conversely, it is seen to be a consequence of propositions already
established. Hence we can inquire, on the one hand, how we have
gradually arrived at a given proposition and, on the other, how we can
finally provide it with the most secure foundation. The first question
may have to be answered differently for different persons; the second
is more definite, and the answer to it is connected with the inner
nature of the propositions considered. The most reliable way of
carrying out a proof, obviously, is to follow pure logic $\ldots$

Everything necessary for a correct inference is expressed in full
$\ldots$ nothing is left to guesswork.
\end{quote}

Hilbert's definition of proof as a sequence of formulae which are
either axioms or generated by rules from elements coming earlier in
the sequence is now quite standard in logic books. Natural Deduction
calculi as introduced by Gentzen (see, e.g.,~\cite{Prawitz65}) are an
extension of this definition. For instance,
Andrews~\cite[p.11]{Andrews86} defines strictly formally the notions
of a proof (from hypotheses and then a proof of a well-formed formula
(wff)). Then he defines \textsl{``A theorem is a wff which has a
  proof.''}

On a more philosophical level there has been a dispute what should
constitute a rigorous proof, since even with this clarification the
question was not fully settled. Most notably there was a dispute
between Hilbert and Brouwer on the right approach to mathematics, in
which the question of constructive proofs versus `classical' proofs
played a major role. Hilbert wanted, in particular, defend the
`paradise' of (infinite) sets provided by Cantor, whereas Brouwer was
wary about the concept of infinity and insisted on
constructiveness. At the time, Brouwer's view was considered by many
mathematicians as too restrictive (and probably is still by many
today). With the advent of computers the idea of constructive proofs,
however, gained great attraction since proving and programming became
the same activity. For details of the dispute
see~\cite{Heijenoort67}. There are other disputes, e.g., about the
axiom of choice and about the rigour of diagrams in reasoning. For the
argument here, it suffices to state that even in the rigorous area of
logical foundations there can be dispute about what should constitute
a proof and what not.\footnote{It should be noted, however, that there
  is a clear language and it is very clear to the participants of a
  dispute what they are talking about.}

\section{The Mathematician's View}\label{sec:maths}
Mathematicians seem to have ignored the development in formal logic to
a large degree.  The start of the rapid development of modern logic can be put
roughly to the mid 19th century. However, the start of the rapid
development of modern mathematics is about 200 years
older.\footnote{And of course there are always precursors, Aristotle
  as the father of logic, Archimedes who was close to inventing
  calculus almost 2000 years before Newton and Leibniz.}

As Kline~\cite[p.256]{Kline80} notes, the ``\textsl{Bourbakists expressed
their position on logic in an article in the \textit{Journal of
  Symbolic Logic\/} (1949): `In other words, logic, so far as we
mathematicians are concerned, is no more and no less than the grammar
of the language which we use, a language which had to exist before the
grammar could be constructed.' Future developments in mathematics may
call for modifications of the logic. This had happened with the
introduction of infinite sets and, $\ldots$ it would happen again.}''

In a similar line, Bourbaki (\cite{Bourbaki54} quoted from
\cite[p.320]{Kline80}) doubts that one of the goals of logicians, to
make mathematics free from contradictions, is feasible:
\begin{quote}\sl
  Historically speaking, it is of course quite untrue that mathematics
  is free from contradictions; non-contradiction appears as a goal to
  be achieved, not as a God-given quality that has been granted to us
  once for all. Since the earliest times, all critical revisions of
  the principles of mathematics as a whole, or of any branch in it,
  have almost invariably followed periods of uncertainty, where
  contradictions did appear and had to be resolved. $\ldots$ There are
  now twenty-five centuries during which the mathematicians have had
  the practice of correcting their errors and thereby seeing their
  science enriched, not impoverished; this gives them the right to
  view the future with serenity.
\end{quote}

Theorems with their proofs are at the core of mathematics and play a
significant role in the working of
mathematicians. Hardy describes in \S~12 of~\cite{Hardy40} two examples
of theorems (with proofs) which he calls `first-rate': First the
theorem that there are infinitely many prime numbers (the proof is
indirect, assume that you have finitely many, multiply them all and
add 1. The new number is not divisible by any prime number, which
gives a contradiction.) and second the theorem that $\sqrt{2}$ is
irrational (again an indirect proof: assume $\sqrt{2}=a/b$ with $a$
and $b$ two integers which have no common divisor. Then $2\cdot
b^2=a^2$. It follows that $a^2$ and hence $a$ must be even, but then
$b$ must be even as well, which gives a contradiction).

In \S~18 Hardy states then: 
\begin{quote}\sl
  In both theorems (and in the
  theorems, of course, I include the proofs) there is a very high
  degree of \textit{unexpectedness}, combined with
  \textit{inevitability} and \textit{economy}. The arguments take so
  odd and surprising a form; the weapons used seem so childishly
  simple when compared with the far-reaching results; but there is no
  escape from the conclusions. There are no complications of detail --
  one line of attack is enough in each case; and this is true too of
  the proofs of many much more difficult theorems, the full
  appreciation of which demands quite a high degree of technical
  proficiency. We do not want many ``variations'' in the proof of a
  mathematical theorem: ``enumeration of cases,'' indeed is one of the
  duller forms of mathematical argument. A mathematical proof should
  resemble a simple and clear-cut constellation, not a scattered
  cluster in the Milky Way.
\end{quote}

Proofs often follow established patterns. Often they are invented,
doubted by other, later generally accepted, and finally re-used,
taught, and generally recognized. Examples are the $\epsilon$-$\delta$
criterion (to establish continuity), diagonalization (to establish the
impossibility of certain properties, e.g.\ halting problem, incompleteness,
uncountability), mathematical induction (to reason about infinite
structures), infinitesimals (to reason about differentiation).

From a logicians point of view mathematical proofs are more like proof
plans.  This is reflected in the education of mathematics. Proof
principles such as the $\epsilon$-$\delta$ criterion are taught in
lectures, without a strictly formal treatment. Many of these
principles are even taught in concrete proofs which have exemplary
character and can be generalized later on to many other cases. Formal
logic, however, is not necessarily part of the education of a
mathematician. In consequence, the concept of a proof is much less
strict, and `mathematics is a {\sc motley} of techniques of proof' as
Wittgenstein put it~\cite[p.~176f]{Wittgenstein84e}.\footnote{Still
  there is a general assumption in mathematics that in principle it is
  possible to extend these mathematical proofs (proof plans) to full
  logic level proofs if necessary.}

This means that the concept of proof is not fixed once and for all but
requires the possibility for extension. Practically, mathematicians
treat proofs and proof methods as first class objects, that is, just
as they introduce new concepts they may introduce new proof
principles, describe them and then use them. For this reason
mathematicians focus on their special fields of expertise and consider
the study of logic as one field among others. And if this field is not
their specialty and particular area of expertise then they do what
professionals do with fields they consider only as marginally
relevant: they give it only marginal attention.

\section{The Deductionist's View}\label{sec:sys}

In formal communities such as the theorem proving community, the
logicians' view of the concept of proof has (for good reasons) been
predominant, but not been the only view. Davis
distinguishes two communities, the logic oriented and the cognitive
oriented communities.

\subsection*{Automated Theorem Proving}

As Davis~\cite[p.5]{Davis01} states: 
\begin{quote}\sl 
  With the ready availability of serious computer power, deductive
  reasoning, especially as embodied in mathematics, presented an ideal
  target for those interested in experiments with computer programs
  that purported to implement the ``higher'' human faculties. This was
  because mathematical reasoning combines objectivity with creativity
  in a way difficult to find in other domains. For this endeavor, two
  paths presented themselves. One was to understand what people do
  when they create proofs and write programs emulating that
  process. The other was to make use of the systematic work of the
  logicians reducing logical reasoning to standard canonical forms on
  which algorithms could be based.
\end{quote}

Since the groundbreaking work in logic in the early 20th century was
very close to implementation it led to the dream to build machines that
can solve hard problems fully automatically. The invention of the
resolution principle by Robinson~\cite{Robinson65} which made search
spaces finitely branching was a great breakthrough and led to the
possibility to prove many theorems fully automatically. In parallel
there was a smaller community which was interested in the cognitive
aspects of theorem proving (by Newell and others, followed up in the
proof planning work by Bundy and others).  At least motivationally the
work is linked to psychological evidence~\cite{Rips94} that deductive
reasoning plays a very important role in human intelligence and that
some proof rules like Modus ponens are universally accepted while
others are accepted only by a minority. Related in this context is
also the work on diagrammatic reasoning (see, e.g.~\cite{Jamnik01})
which shows that reasoning falsely considered for some time as
inferior, can be made very precise.

In general, however, the dream of full automation has not come true at
large. There are fascinating exceptions such as the proof of the
Robbins problem (\cite{McCune97}), but still mathematicians do not
have theorem proving machines on their desks which they use to a
similar degree as they use typesetting programs or computer algebra
systems.  And possibly not everybody would want such a machine, since
as Hardy put it in~\cite[\S~10]{Hardy40} ``\textsl{there is nothing in
  the world which pleases even famous men $\ldots$ quite so much as to
  discover, or rediscover, a genuine mathematical theorem.}'' and we
can add ``\textsl{and a genuine mathematical proof.}'' (but proofs are
parts of the theorem for Hardy anyway.) Why leaving the fun to a
machine?

There has been a different community at least since the 1960s, namely
the community interested in being able to check proofs by a
machine. On first sight this looks like a much less ambitious goal,
but it turned out to be actually much more difficult than
anticipated. We will look at this next.

\subsection*{Proof Checking}

The perhaps two most prominent approaches to proof checking -- from
which other systems have been derived -- are Automath~\cite{Bruijn80}
and Mizar~\cite{Trybulec80}. The goal is not to find proofs
automatically but to check proofs.  De Bruijn summarizes his dream in
1994~\cite[p.210]{Nederpelt94} as follows:

\begin{quote}\sl
As a kind of dream I played (in 1968) with the idea of a future where
every mathematician would have a machine on his desk, all for himself,
on which he would write mathematics and which would verify his work.
But, by lack of experience in such matters, I expected that such
machines would be available in 5 years from then. But now, 23 years
later, we are not that far yet.\\\hspace*{\fill}
\end{quote}

In many ways this dream is more exciting, since firstly it looks more
feasible and secondly it is something mathematicians and professionals
working in related fields can appreciate more. Although proof checkers
have been extended by useful extensions which allow for higher-level
proofs, most notably by tactics which allow to reduce many steps in a
proof to a single user interaction, even 16 years after de Bruijn's
retrospective we are still not there and mathematicians do not widely
use the corresponding systems. However, they can be used and some do
use them, most notably there is the Flyspeck Project~\cite{Hales10} in
which Hales (and others) are formalizing his proof of the Kepler
conjecture.

\section{How to Make Systems more Accepted?}

Systems which deal with proofs can be built for different purposes and
different purposes result in different requirements. Only some of them
are currently adequately supported by mathematical knowledge
management systems. Let us look at the most common purposes/contexts
in which proofs are communicated:

\begin{description}
\item[Education:] In an educational context proofs will be presented
  and/or jointly developed in order to teach the concept. A teacher
  may want to teach how to find a proof but more typically will teach
  how to write up a proof so that it is of an acceptable
  standard. These are two different modes as
  P{\'o}lya~\cite[p.~vi]{Polya54}, pointed out:
\begin{quote}\sl
  We secure our mathematical knowledge by {\it demonstrative
    reasoning}, but we support our conjectures by {\it plausible
    reasoning} $\ldots$ Demonstrative reasoning is safe, beyond
  controversy, and final.  Plausible reasoning is hazardous,
  controversial, and provisional.  $\ldots$ In strict reasoning the
  principal thing is to distinguish a proof from a guess, a valid
  demonstration from an invalid attempt.  In plausible reasoning the
  principal thing is to distinguish a guess from a guess, a more
  reasonable guess from a less reasonable guess.  $\ldots$ [plausible
  reasoning] is the kind of reasoning on which his [a mathematician's]
  creative work will depend.
\end{quote}

\item[Proof development:] Here the scenario is that of a mathematician
  or a group of mathematicians developing a proof. They do not know
  yet the details of the proof (or even whether there is a proof),
  they may have some ideas which may be vague and informal. A
  blackboard and a piece of chalk seem to be the tools of choice and
  systems at best offer the functionality of a blackboard and
  chalk. In P\'olya's words, the game is mainly about plausible reasoning at
  this stage. The task of providing support is particularly
  challenging since proof attempts, ideas, partial proof plans may have
  to be communicated.

\item[Automation:] If automation is the objective and proofs are
  found, automated theorem provers typically can document a formal
  proof object which can be independently checked. This object can be
  communicated.

\item[Correctness:] If correctness of arguments is sought then proofs
  must be checkable. At a calculus level the different formal systems
  implemented allow this. Human mathematicians, however, can check
  proofs at a less formal level. Support at this level is still
  patchy, although important steps have been made in an area which is
  labelled as the development of a mathematical vernacular (going back
  to de Bruijn and the Automath project, and continued by Nederpelt
  and Kamareddine~\cite{Nederpelt01}).

\end{description}

For any of the different activities there is the question: What kind
of information is necessary and how should it be represented?

A proof is an argument that should convince the reader (interpreter)
of the truth of a statement (certain axioms and assumptions
given). That is, a proof is a relationship between the argument and
the reader, and the reader has to come with some level of knowledge.

If we know a lot, then a proof can be more concise. If we know the
theorem already then we do not need to be convinced. If we know
little, then we need a detailed argument which convinces us beyond
reasonable doubt (some may say beyond any doubt) of the correctness of
the theorem. In this respect a proof is a proof only with respect to a
receiver/reader.  ``\textsl{Nothing can be explained to a stone, the
  reader must understand something beforehand.}'' as McCarthy
formulated it (1964, p.7), quoted from~\cite[p.8]{Abrams70}
and analogously we can state that ``\textsl{Nothing can be
  explained to God, since he understands everything beforehand.}'' or
as Ayer~\cite[p.85f]{Ayer36} put it:
\begin{quote}\sl
The power of logic and mathematics to surprise us depends, like their
usefulness, on the limitations of our reason. A being whose intellect
was infinitely powerful would take no interest in logic and
mathematics. For he would be able to see at a glance everything that
his definitions implied, and, accordingly could never learn anything
from logical inference which he was not fully conscious of
already. But our intellects are not of this order. It is only a minute
proportion of the consequences of our definitions that we are able to
detect at a glance. Even so simple a tautology as ``\,$91\times 79 =
7189$'' is beyond the scope of our immediate apprehension. To
assure ourselves that ``7189'' is synonymous with ``\,$91\times
79$'' we have to resort to calculation, which is simply a process
of tautological transformation -- that is, a process by which we
change the form of expression without altering their significance. The
multiplication tables are rules for carrying out this process in
arithmetic, just as the laws of logic are rules for the tautological
transformation of sentences expressed in logical symbolism or in
ordinary language.
\end{quote}

Typically, we are in between the stone and God: We know certain
theorems and proofs and are happy to accept certain arguments when
they are mentioned in a new proof and others not. We can fill in certain
gaps, but not others. We have intelligence which goes beyond checking
substitutions and matching, which can convince us that a theorem is
really a theorem. A proof should give us a good reason why we should
not doubt the correctness of the theorem at an appropriate
level. Going back to a logic level proof is typically like being dragged on a
level on which we do not see the wood for the trees. 

Indeed proofs come in various formats, they can be presented at
different levels of abstraction and can be quite different in style
and details. In order to represent and support them appropriately we
need to know what they are needed for and have to reflect the purpose
and the level of understanding and knowledge of the reader. The reader may
know the proof already or know a similar proof (and would be quite quick
at understanding the new one). The reader may have no intuition -- possibly the
statement is even counter intuitive -- and would have to check steps
slowly and carefully. Or the reader may not be able to understand the proof
in a reasonable amount of time at all since they lack the corresponding
knowledge and would require a significant course in a whole field of
mathematics before they can appreciate the
arguments.\footnote{Obviously the borders are not sharp. We may know a
  similar proof, and actually we would not remember every single
  step. Having a good intuition, having some intuition, and having no
  intuition, or a counter intuition is again fluid.  The proof of the
  Robbins problem was so hard for humans since they did not have an
  intuition of the Robbins algebras.}

In a familiar area, mathematicians know which arguments to accept and
where to be careful. They are well aware of fallacies to avoid, that
is, we have positive and negative information at our disposal and
avoid the fallacies as they are described by Maxwell
in~\cite{Maxwell59}. Maxwell's examples deal, for instance, with
non-apparent divisions by zero, with problems with integration by
parts, and with incorrectly drawn auxiliary diagrams in geometric
proofs. Maxwell distinguishes between fallacies, where things go wrong
on a deeper level (and proof checking on a high-level may wrongly
succeed) and howlers, where the incorrectness of the argument is
apparent (and a wrong argument may still give the correct result).

That the mathematical notion of a proof is subject to change, not
strictly formal, and not beyond doubt has most convincingly been
described by Lakatos~\cite{Lakatos76} in an analysis of the history of
the Euler polyhedron theorem, which had an exciting history of proofs
and subsequent counterexamples, which led to improved proofs and more
sophisticated counterexamples. This has not led to a general distrust
in proofs. Although the theorem is not central to mathematics, still
-- as Hardy put it \cite[\S~12]{Hardy40} -- \textsl{a mathematician
  offers} \textit{the game}, and a contradiction may cast doubt on the
correctness of mathematics as a whole. However, the sequence of proof,
counterexample, proof can be seen very much in the spirit of the quote
in section~\ref{sec:maths} of Bourbaki that ``\textsl{mathematicians
  have had the practice of correcting their errors and thereby seeing
  their science enriched, not impoverished.}''

Theorem proving and checking proofs is a social activity and in a
highly specialized society there are different reasons why we believe
a theorem and its proof. Only few will actually have the knowledge,
the capacity, and the time to understand complicated proofs like that
of the Fermat-Wiles theorem or the Kepler conjecture. Still most of us
will accept that there are proofs and that the theorems hold. The two
theorems mentioned by Hardy, however, have much simpler proofs and it
belongs to the folklore to know their proofs.

We see that there is a broad spectrum of proofs. Typically natural
language in combination with diagrams is used to store and communicate
proofs. Some types of proof (formal logical proofs, some types of
proof plans) can be represented in a format which is better suited to
mechanical manipulation (e.g.\ to proof checking) than natural
language. Other types are still difficult to formalize. Work on the
mathematical vernacular is certainly useful in order to formalize the
variety of proofs. An advanced approach to understand informal proofs
at a linguistic level has been carried through by
Zinn~\cite{Zinn04}. He analyzes the linguistic structure of proofs and
builds internal structures, which reflect the inner logic of the
proofs. This opens a way to understanding and checking informal
mathematical discourse.

\section{Summary}\label{sec:summary}

Proofs come at different levels and with different intentions. They
are written for readers/checkers who/which must have certain
competences. A human mathematician who knows a theorem very well knows
and can communicate proofs of it at different levels: the gist of it,
which allows other experts to reconstruct a full proof, a proof plan
for a less proficient reader/checker, and a low level proof for a
checker with little information in the field. Likewise an expert can
understand proofs on different levels.

A system that has deep knowledge about proofs would be able to link
the different levels.  Achieving such a human level of expertise looks
AI-hard unfortunately. On the other hand this has its attraction as
Davis states, since it ``\textsl{combines objectivity with
  creativity}.''  (Generalized) proof plans can offer a framework
which is general enough to capture the different levels. Linking
different levels and understanding different levels simultaneously
will remain a hard problem for some time, and proof will remain a
colourful concept.


\begin{thebibliography}{10}

\bibitem{Abrams70}
Philip~S. Abrams.
\newblock An {APL} machine.
\newblock SLAC-114 UC-32 (MISC), Stanford University, Stanford, California,
  1970.

\bibitem{Andrews86}
Peter~B.\ Andrews.
\newblock {\em An Introduction to Mathematical Logic and Type Theory: To Truth
  through Proof}.
\newblock Academic Press, Orlando, Florida, USA, 1986.

\bibitem{Ayer36}
Alfred~Jules Ayer.
\newblock {\em Language, Truth and Logic}.
\newblock Victor Gollancz Ltd, London, United Kingdom, 2nd edition, 1951 edition, 1936.

\bibitem{Bourbaki54}
Nicolas Bourbaki.
\newblock {\em {T}h\'eorie {d}es {e}nsembles}.
\newblock {\'E}l\'ements de math\'ematique, Fascicule 1. Hermann, Paris,
  France, 1954.

\bibitem{Bruijn80}
Nicolaas Govert~de Bruijn.
\newblock A survey of the project Automath.
\newblock In J.P. Seldin, J.R. Hindley, editors, {\em To H.B.~Curry - Essays
  on Combinatory Logic, Lambda Calculus and Formalism}, pages 579--606.
  Academic Press, London, United Kingdom, 1980.

\bibitem{Davis01}
Martin Davis.
\newblock The early history of automated deduction.
\newblock In Alan Robinson and Andrei Voronkov, editors, {\em Handbook of
  Automated Reasoning -- Volume I}, pages 5--14. Elsevier Science, Amsterdam,
  The Netherlands, 2001.

\bibitem{Frege79}
Gottlieb Frege.
\newblock {B}egriffsschrift, {e}ine {d}er {a}rithmetischen {n}achgebildete
  {F}ormelsprache {d}es {r}einen {D}enkens.
\newblock Halle, 1879.

\bibitem{Hales10}
Thomas Hales.
\newblock The {F}lyspek {P}roject.
\newblock \url{http://code.google.com/flyspeck/}, 2010.

\bibitem{Hardy40}
Godfrey Hardy.
\newblock {\em A Mathematician's Apology}.
\newblock Cambridge University Press, London, United Kingdom, 1940.

\bibitem{Heijenoort67}
Jean~van Heijenoort, editor.
\newblock {\em From Frege to G{\"o}del -- A Source Book in Mathematical Logic,
  1879-1931}.
\newblock Harvard Univ.~Press, Cambridge, Massachusetts, USA, 1967.

\bibitem{Jamnik01}
Mateja Jamnik.
\newblock {\em Mathematical Reasoning with Diagrams: From Intuitions to
  Automation}.
\newblock CSLI Press, Stanford, California, USA, 2001.

\bibitem{Kline80}
Morris Kline.
\newblock {\em Mathematics -- The Loss of Certainty}.
\newblock Oxford University Press, New York, USA, 1980.

\bibitem{Lakatos76}
Imre Lakatos.
\newblock {\em Proofs and Refutations}.
\newblock Cambridge University Press, 1976.

\bibitem{Maxwell59}
E.A. Maxwell.
\newblock {\em Fallacies in Mathematics}.
\newblock Cambridge University Press, Cambridge, United Kingdom, 1959.

\bibitem{McCune97}
William McCune.
\newblock Solution of the {R}obbins problems.
\newblock {\em Journal of Automated Reasoning}, {\bf 19}(3):263--276, 1997.
\newblock See also \url{http://www.mcs.anl.gov/home/mccune/ar/robbins/}.

\bibitem{Nederpelt94}
Rob Nederpelt, Herman Geuvers, and Roel de~Vrijer, editors.
\newblock {\em Selected Papers on Automath}, volume 133 of {\em Studies in
  Logic and the Foundations of Mathematics}.
\newblock North-Holland, Amsterdam, The Netherlands, 1994.

\bibitem{Nederpelt01}
Rob Nederpelt and Fairouz Kamareddine.
\newblock An abstract syntax for a formal language of mathematics.
\newblock In {\em The Fourth International Tbilisi Symposium on Language, Logic
  and Computation}, 2001.
\newblock
  \url{http://www.cedar-forest.org/forest/papers/conference-publications/tbilisi01.ps}.

\bibitem{Polya54}
George P{\'o}lya.
\newblock {\em Mathematics and Plausible Reasoning}.
\newblock Princeton University Press, Princeton, New Jersey, USA, 1954.
\newblock {Two volumes, Vol.~1}:
  Induction and Analogy in Mathematics, Vol.~2:
  Patterns of Plausible Inference.

\bibitem{Prawitz65}
Dag Prawitz.
\newblock {\em Natural Deduction -- A Proof Theoretical Study}.
\newblock Almqvist \& Wiksell, Stockholm, Sweden, 1965.

\bibitem{Rips94}
Lance~J.\ Rips.
\newblock {\em The Psychology of Proof -- Deductive Reasoning in Human
  Thinking}.
\newblock The MIT Press, Cambridge, Massachusetts, USA, 1994.

\bibitem{Robinson65}
John~Alan Robinson.
\newblock A machine oriented logic based on the resolution principle.
\newblock {\em Journal of the ACM}, {\bf 12}:23--41, 1965.

\bibitem{Trybulec80}
Andrzej Trybulec.
\newblock The {M}izar logic information language.
\newblock {\em Studies in Logic, Grammar and Rhetoric}, {\bf 1}, 1980.
\newblock Bia{\l}ystok, Poland.

\bibitem{Whitehead10}
Alfred~North Whitehead and Bertrand Russell.
\newblock {\em Principia Mathematica}, volume~I.
\newblock Cambridge University Press, Cambridge, United Kingdom; 1910.

\bibitem{Wittgenstein84e}
Ludwig Wittgenstein.
\newblock {\em {B}emerkungen {\"u}ber {d}ie {G}rundlagen {d}er {M}athematik},
  volume 506.
\newblock Suhrkamp-Taschenbuch Wissenschaft, Frankfurt, Germany; 3rd edition,
  1989.

\bibitem{Zinn04}
Claus Zinn.
\newblock {\em Understanding Informal Mathematical Discourse}.
\newblock PhD thesis, Friedrich-Alexander-Universit\"at Erlangen-N\"urnberg,
  Erlangen, Germany, 2004.

\end{thebibliography}
\end{document}